\documentclass[11pt]{article}
\usepackage{graphicx}
\usepackage[boxruled,vlined,linesnumbered]{algorithm2e}
\usepackage{amsbsy}
\usepackage{amssymb}
\begin{document}

%\begin{frontmatter}

\title{A Discrete Firefly Algorithm to Solve a Rich Vehicle Routing Problem Modelling a Newspaper Distribution System with Recycling Policy}
%% \titlerunning{A DFA to Solve an R-VRP Modelling a Newspaper Distribution System}

\author{Eneko Osaba$^{1,2}$,         
				Xin-She Yang$^1$, 			
				F. Diaz$^2$,       
				E. Onieva$^2$, 
				A. D. Masegosa$^2$ ,  
				A. Perallos$^2$ \\[10pt]
1)         E. Osaba, X.S. Yang \\
              School of Science and Technology, Middlesex University\\
							Hendon Campus, London, NW4 4BT, United Kingdom\\	
2) E. Osaba, F. Diaz, E. Onieva, A. Masegosa, A. Perallos \\
              Deusto Institute of Technology (DeustoTech), University of Deusto\\
              Av. Universidades 24, Bilbao 48007, Spain\\
}

\date{\hrule{\bf Citation Details}: \\
 E. Osaba, Xin-She Yang, F. Diaz, E. Onieva, A. D. Masegosa, A. Perallos,  \\
A Discrete Firefly Algorithm to Solve a Rich Vehicle Routing Problem Modelling a Newspaper Distribution System with Recycling Policy, {\it Soft Computing}, Published First Online 18 March 2016. DOI 10.1007/s00500-016-2114-1
\hrule }

\maketitle

\begin{abstract}
A real-world newspaper distribution problem with recycling policy is tackled in this work. In order to meet all the complex restrictions contained in such a problem, it has been modeled as a rich vehicle routing problem, which can be more specifically considered as an asymmetric and clustered vehicle routing problem with simultaneous pickup and deliveries, variable costs and forbidden paths (AC-VRP-SPDVCFP). This is the first study of such a problem in the literature. For this reason, a benchmark composed by 15 instances has been also proposed. In the design of this benchmark, real geographical positions have been used, located in the province of Bizkaia, Spain. For the proper treatment of this AC-VRP-SPDVCFP, a discrete firefly algorithm (DFA) has been developed. This application is the first application of the firefly algorithm to any rich vehicle routing problem. To prove that the proposed DFA is a promising technique, its performance has been compared with two other well-known techniques: an evolutionary algorithm and an evolutionary simulated annealing. Our results have shown that the DFA has outperformed these two classic meta-heuristics.

\end{abstract}

\section{Introduction}
\label{sec:intro}

Transportation is an important factor for today's smart society. Public transportation, for example, is used by almost the whole population, and it directly affects the quality of life. However, modelling and planning such complex transport system is a very challenging task. Here, this paper focuses on another sort of transport: the transportation in the business world. Due to the rapid advancement of technologies, logistics systems have become very important for media companies. The fact that anyone in the world can be well connected has led to complex transport networks that are very demanding and are becoming increasingly important. Therefore, an efficient logistic network can make a huge difference for companies and relevant business operations.

To cite one fact that highlights the importance of logistics in this sector, in some businesses, such as groceries delivery, the distribution costs can lead to an increase in the product price up to 70\% \cite{golden1987or}. Thanks to cases like this, it is obvious to show the importance of this sector.

Therefore, this paper focuses on a real-world logistic problem, and its effective treatment. The real-world situation dealt in this work is related to the daily delivery of newspapers. Specifically, the object of this study is a medium-sized newspaper distribution company, which offers its services at the regional level. This company has certain mandatory principles, which must be taken into account when performing the daily planning tasks of deliveries. One of these principles is a strict recycling system. Another one is the treatment of the different towns, or cities, as separate units. Besides that, the company considers certain factors for the scheduling process, as variable travel times (depending on the hour they are carried out), or the transit prohibition in certain streets. Although this paper is focused on a specific company located in Bizkaia (Spain), it is noteworthy that the goal of this work is to develop a model that is applicable to every company of similar characteristics.

Therefore, the aim of this paper is to address efficiently this Newspaper Distribution System with Recycling Policy (NDSRP). For this purpose, the NDSRP has been modeled as a Multi-Attribute, or Rich-Vehicle Routing Problem (R-VRP). Nowadays, as indicated in \cite{MAVRP} or \cite{lahyani2015rich}, R-VRPs form a hot topic in the scientific community. These kinds of problems are special cases of the well-known conventional vehicle routing problem (VRP) \cite{toth2001vehicle}, with the distinction of having multiple constraints and complex formulations. This sort of problems can have a great scientific interest, since such NP-Hard problems present a tough challenge to solve. Furthermore, their social interest is also high, as their applicability to real-world situations is greater than the conventional versions of routing problems.

To be more specific in the present paper, an Asymmetric and Clustered Vehicle Routing Problem with Simultaneous Pickup and Deliveries, Variable Costs and Forbidden Paths (AC-VRP-SPDVCFP) was proposed to address the presented NDSRP. Some examples of recently developed R-VRPs can be \cite{de2015gvns} or \cite{amorim2014rich}. In the former work, the authors propose an R-VRP with hard and soft time windows, heterogeneous fleet, customers priorities and vehicle-customer constraints. The solution proposed by the authors of this paper has already been integrated into the optimization tool of a fleet management system used in the Canary Islands. In the latter work, an R-VRP was proposed to deal with the perishable food management. In this work, a heterogeneous fleet site-dependent VRP with multiple time windows was presented. Another example of recently developed R-VRP can be the proposed in \cite{lahyani2015multi}. In their paper, an R-VRP was developed for the olive oil collection in Tunisia. The R-VRP designed in this work was a multi-product, multi-period and multi-compartment VRP. These were some of the examples that justified the increasing interest of R-VRP in the scientific community. For further information on R-VRP, readers can refer to the surveys \cite{MAVRP} and \cite{lahyani2015rich}.

Though some appropriate methods can be found in the literature to address such complex problems,  arguably the most successful techniques to solve these R-VRPs are the heuristics and metaheuristics \cite{caceres2014rich}. In this study, the attention has been focused in the last ones. In fact, to solve the AC-VRP-SPDVCFP proposed in this paper one metaheuristic has been developed. Some classical examples of metaheuristics are the Tabu search \cite{tabuSearch} and simulated annealing (SA) \cite{simulatedAnnealing} as local search methods, and genetic algorithm (GA) \cite{genetic1,genetic2}, ant colony optimization \cite{ACO} and particle swarm optimization \cite{PSO} as population ones.
Though such methods were proposed some decades ago, they still attract the attention in the scientific community \cite{rodriguez2015rwa,cao2015tabu,inkaya2015ant} and metaheuristics, especially new metaheuristic algorithms and their proper implementations still form a hot topic in the field. In fact, many different metaheuristics have been proposed in the last two decades, which have been successfully applied to various problems and fields. Some examples of these techniques are the imperialist competitive algorithm, presented in 2007 by Gargari and Lucas \cite{imperialist}, the bat algorithm, proposed by Yang in 2010 \cite{yang2010new}, or the harmony search, presented in 2001 by Geem et al. \cite{geem2001new}.

Another type of technique that shows a good performance applied to routing problems are the memetic algorithms \cite{moscato2003gentle,nalepa2014co}. In \cite{vidal2013hybrid}, for example, a hybrid genetic algorithm is presented for solving a large class of vehicle routing problems with time-windows. On the other hand, in \cite{vidal2012hybrid} a hybrid genetic algorithm is presented to solve an R-VRP composed by multidepot and periodicity features. In \cite{qi2015decomposition} a decomposition based memetic algorithm is proposed for a multi-objective VRP with time windows. Another kind of R-VRP is solved by a hybrid techniques in \cite{bortfeldt2015hybrid}. In this case, the characteristics of the problem are clustered backhauls and 3D loading constraints. Finally, a multiperior VRP with profit is addressed by a memetic algorithm in \cite{zhang2013memetic}. Works cited in this paragraph are some recent and interesting examples of the whole literature, some other recent examples can be found in, for example, \cite{nagata2010penalty,nalepa2015adaptive,marinakis2015memetic}.

Therefore, in this paper one metaheuristic proposed a few years ago is used to solve the presented problem. This technique is the firefly algorithm (FA), proposed by Yang in 2008 \cite{yang2008nature}. This nature-inspired algorithm is based on the flashing behaviour of fireflies, which acts as a signal system to attract other fireflies. As can be seen in several surveys \cite{FS2,FS3}, the FA has been successfully applied to many different optimization fields and problems since its proposal. In addition, it still attracts a lot of interests in the current scientific community \cite{ma2015navigability,liang2015enhanced,zouache2015quantum}. Nevertheless, the FA has never been applied to any R-VRP. This lack of works, along with the growing scientific interest in bio-inspired algorithms, and the good performance shown by the FA since its proposal in 2008, has motivated its use in this study.

In this way, the main novelties and contributions of this paper can be listed in the following way. It is noteworthy that these originalities and contributions have motivated the realization of this work:

\begin{itemize}
	\item In this paper, a DPRP is dealt with using an R-VRP. As will be mentioned in the following section, the problem of newspaper distribution has been previously addressed in the literature, but never using an R-VRP as complex as the one presented in this paper. Furthermore, it is the first time that an R-VRP of this characteristics is proposed in the literature. For this reason, a benchmark composed by 15 instances has been developed, based on real geographic locations.

	\item In this work, a discrete version of the FA (DFA) is presented for solving the proposed R-VRP. Until the time of writing, the FA has never been applied to any R-VRP. Anyway, the main novelty of the presented DFA is not only its application field. The technique developed in this work uses the Hamming Distance function to measure the distance between two fireflies of the swarm. This approach has been rarely used previously, as well as the move strategy used by the fireflies, which is based on evolution strategies. All these characteristics are described in following sections. Additionally, in order to prove that the DFA is a promising technique to solve the designed AC-VRP-SPDVCFP, the results obtained by the DFA are compared with the ones obtained by the evolutionary algorithm (EA) based on mutations and the evolutionary simulated annealing (ESA) \cite{ESA}.
	
\end{itemize}

The rest of the paper is organized as follows. In Section \ref{sec:RealProblem}, the real-world problem addressed in this work is described. In Section \ref{sec:VRP}, the R-VRP proposed to deal with the real problem is described in detail. Furthermore, in Section \ref{sec:firefly}, the developed DFA is described in detail. After that, the experimentation carried out is detailed in Section \ref{sec:exp}, as well as the benchmark designed for the presented R-VRP. This paper then concludes with some brief conclusions and further research topics in Section \ref{sec:conc}.

\section{Newspapers Distribution with Recycling Policy}
\label{sec:RealProblem}

As has been pointed in the previous section, the real-world situation faced in this work is related to the newspaper distribution. More specifically, the object of study is a medium-sized newspaper distribution company. The area of the coverage of this company is at a provincial level. This means that the company has to serve a set of customers geographically distributed in separate towns and cities. The company in question has some principles, which are the base of their logistics planning. The first principle is to treat towns, or cities, as separate units. In this way, if one vehicle enters a city, or a town, it is forced to serve each and every customer located therein. Therefore, one vehicle is banned for entering one city or town whether it does not have sufficient capacity to meet the demand of all customers deployed there.

On the other hand, due to the current environmental requirements, the company has a simple but robust paper recycling policy. In this case, the objects to recycle are the newspapers that were not sold the previous day. Thus, as can be deduced, vehicles not only have to meet the delivery demands of the customers. Besides that, they have to collect at each point those newspapers that have not been sold the day before.

In addition,  the company has to take into account certain factors in the routes planning process. The first one is related to the hours at which the deliveries and collections are done. The service is performed daily during morning from 6:00am to 15:00. Within this time window exists one range, established between 8:00am and 10:00am, considered as ``peak hours''. In this way, traveling costs from one point to another are greater if they are performed at ``peak hours''. Besides this, with the aim of respecting all the traffic rules, vehicles cannot go through prohibited roads.

Throughout the past few decades, the problem of the newspaper delivery has been addressed many times in the literature. In 1996, Ree and Yoon presented a two-stage heuristic for a newspaper distribution problem, taking as a case study of a major press company of Korea \cite{ree1996two}. The problem used in this study is a multi-VRP with time windows, which is faced in two different stages by the proposed heuristic. On the other hand, in \cite{archetti2013heuristic} a free newspaper delivery problem is addressed. In their study, the problems was decomposed in two phases. In the first stage, the delivery plan was created, based on concept taken from the Inventory Routing Problem \cite{campbell1998inventory}. Once the delivery plan was fixed, the resulting problem was solved in the second stage as a variant of the conventional VRP with time windows (VRPTW) \cite{kallehauge2005vehicle}. They used a heuristic for solving this second stage problem. The same VRPTW was used in \cite{boonkleaw2009strategic} to face a case study of the newspaper delivery problem focused in the city of Bangkok and a modified sweep algorithm was presented to solve the proposed VRPTW. These papers are some of the examples that can be found in the literature. Many others can be found, considering, for example, the problem of delivering newspapers to private subscribers \cite{hurter1996newspaper,van1999solving}.

Regarding the problem addressed in this paper, some originalities and novelties are proposed: the features of the variable travel times and forbidden paths have never been used for the newspaper distribution system, as well as the recycling policy applied in this work. In addition, as we will see
in the rest of the paper, the R-VRP proposed in this work has a great number of constraints, making easier its application to the real world. Finally, the technique developed in this work solves the proposed AC-VRP-SPDVCFP in only one phase, in contrast with the approaches presented in \cite{ree1996two,archetti2013heuristic}. This is an advantage because solving the problem in only one phase, the runtimes and the computational effort can be considerably reduced.

\section{Asymmetric and Clustered Vehicle Routing Problem with Simultaneous Pickups and Deliveries, Variable Costs and Forbidden Paths}
\label{sec:VRP}

As has been stated earlier in this paper, the real-world situation of the newspaper distribution has been modeled as an R-VRP. In this section, a detailed description of the presented problem is provided. First of all, in Section \ref{sec:generalDesc}, the basic features of the problem are detailed, followed by the mathematical formulation of the proposed R-VRP is depicted in Section \ref{sec:mathDesc}.

\subsection{General description of the problem}
\label{sec:generalDesc}

The proposed R-VRP has the following general characteristics. It is noteworthy that all these features proposed here are to take into account the conditions stated in Section \ref{sec:RealProblem}. Besides those, some additional restrictions have been considered with the intention of developing a model closer to real conditions.

\begin{enumerate}
		
	\item \textit{Asymmetry}: The traveling costs in the proposed R-VRP are asymmetric. This means that the traveling cost from any $i$ node to another $j$ node is different from the reverse trip cost. It is important to highlight that, in the problem proposed in this paper, this asymmetry appears in every node-to-node travel. This feature is not common in most routing problems that can be found in the literature, and it brings realism to the problem. Anyway, asymmetric costs have been applied previously in the literature in a different context \cite{assy,toth1999heuristic,herrero2014solving}. It is noteworthy that this feature is very valuable in real-world applications.
	
	\item \textit{Clusterized}: With this attribute, the different clients that make up the system are grouped in different sets or clusters. In this case, each cluster represents a city or town. The condition that vehicles must meet is the following: if one vehicle meets the demand of any customer belonging to any cluster, this vehicle must serve each and every customer of this cluster. In this way, one vehicle cannot enter a cluster if it does not have enough capacity to serve all customers in this city or town. This feature has been used previously in many studies in a different context. \cite{clustered1,clustered2,vidal2014hybrid}.
			
	\item \textit{Simultaneous Pickup and Delivery}: This property is an adaptation of the often used pickup and delivery system of some routing problems \cite{wang2015parallel,li2015iterated}. Basically, this system consists in the existence of two types of nodes: the \textit{delivery nodes} and the \textit{collect nodes}. The former ones are those points in where newspapers are delivered. On the other hand, in \textit{collect nodes}, newspapers are collected, with the intention of bringing them back to the center for their subsequent recycling.
	
In addition, it is important to highlight that, due to the simultaneous nature of the real-world situation, one customer can ask for both delivery and collection of newspapers. In this way, both \textit{delivery nodes} and \textit{delivery-collect nodes} can be found in the system. On the other hand, it is assumed that all customers request the delivery of newspapers. For these reason, customers which demand only the collection of newspapers have not been taken into account.
	
	\item \textit{Variable Travel Times}: In real transportation situations, the travel between two points does not always take the same cost, either temporarily or economically. In many cases, this cost is subject to some external variables, such as the hour, the traffic or the weather. With the intention of adding more realism to the problem, this situation has been represented in the R-VRP proposed in this work. To this end, it has established a schedule between 6:00am and 15:00. Within this schedule, two different time-periods have been established, called ``peak hours" and ``off-peak hours". Peak hours are composed by two hours, between 8:00am and 10:00am. All trips performed in this time window imply a higher cost, comparing with the costs of conducting the same trip in ``off-peak" period. A similar characteristic has been previously used a few times in the literature \cite{variableCosts}.
	
	\item \textit{Forbidden Paths}: In the real world, it is common to find one-way roads, where the traffic in a particular direction is prohibited. There are also pedestrian streets, where vehicles cannot go through. With the aim of recreating this common situation, the problem has certain arcs $(i,j)$ which cannot be used in the final solution. A similar philosophy has been used previously in some studies in the literature \cite{forbidden}.
\end{enumerate}

With the above assumptions and simplifications, the proposed AC-VRP-SPDVCFP is an R-VRP, whose objective is to find a set of $r$ routes, trying to minimize the total cost of the complete solution, and taking into account the two different kinds of nodes, respecting the restrictions of the clusters and vehicles capacities ($Q$) and not traveling through any forbidden path. As can be seen, being an R-VRP problem, the AC-VRP-SPDVCFP has multiple constraints. It is important to point out that all these restrictions reduce the size of the search space which encompasses the feasible solutions. Anyway, all these constraints make the process of generating feasible solutions and successors to be a very complex task.

\begin{figure*}[tb]
	\centering
		\includegraphics[width=0.6\textwidth]{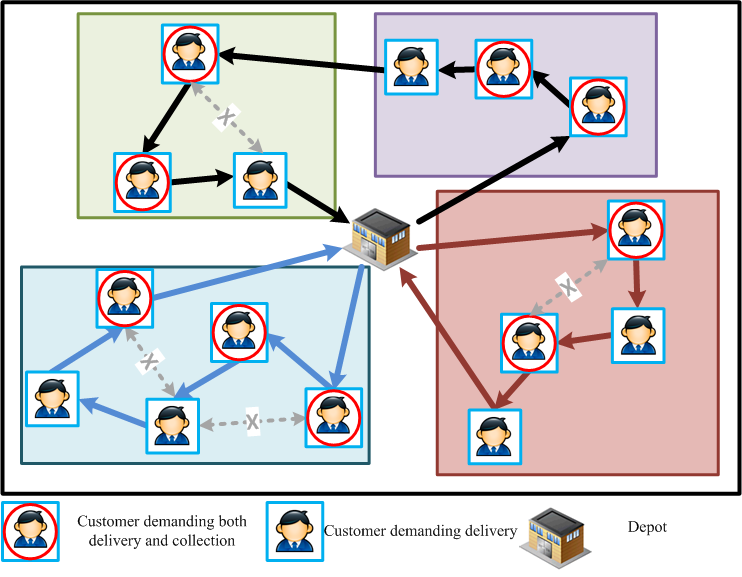}
	\caption{Possible AC-VRP-SPDVCFP instance composed by 14 nodes, and one feasible solution. Gray paths represent forbidden arcs.}
	\label{fig:ACVRPSPDVCFP}
\end{figure*}

In Figure \ref{fig:ACVRPSPDVCFP}, a possible 14-noded instance of the proposed AC-VRP-SPDVCFP is represented. A feasible solution for this instance is also shown in Figure \ref{fig:ACVRPSPDVCFP}.

\subsection{Mathematical description of the problem}
\label{sec:mathDesc}

The presented AC-VRP-SPDVCFP can be defined on a complete graph $G= (V,A)$ where $V=\{v_0,v_1,\dots,v_n\}$ is the set of vertices which represent the nodes of the system. On the other hand, $A= \{(v_i,v_j): v_i,v_j  \in V,i\neq j \}$ is the set of arcs which represent the interconnections between nodes. Each arc has an associated cost $d_{ij}$. Due to the asymmetry feature,  $d_{ij} \neq d_{ji}$. It is noteworthy that, in this case, the cost of traveling from $i$ to $j$ is always different from the cost of traveling from $j$ to $i$. The cost of traveling via a forbidden arc is infinite. In this way, it is ensured that they will not appear in the final solution. Furthermore, the vertex $v_0$ represents the depot, and the rest are the visiting points. Additionally, $V$ is divided into $c+1$ mutually exclusive nonempty subsets, $C=\{V_0,V_1,...,V_c\}$, each one for each cluster. These subsets hold the following conditions:
\begin{equation}
	 V = V_0 \cup V_1 \cup ... \cup V_c \label{clustered1}
\end{equation}
\begin{equation}
	 V_a \cap V_b = \emptyset, \ \ \ a,b \in {0,1,...,c}, a \neq b \label{clustered2}
\end{equation}

It is noteworthy that $V_0$ only contains $v_0$, which depicts the depot. The remaining $n$ nodes are divided into $c$ clusters. Besides that, each node $i$ has assigned two kinds of demands: one associated with the delivery $d_i$ $>$ 0 and the other with the pick-up $p_i$ $\geq$ 0.

The proposed AC-VRP-SPDVCFP can be mathematically formulated in the following way. It is important to highlight that $y_{ij}$ depicts the demand picked-up in clients routed up to node $i$ (including node $i$) and transported in arc $(i,j)$. On the other hand, $z_{ij}$ represents the demand to be delivered to clients routed after node $i$ and transported in arc $(i,j)$ \cite{montane2006tabu}. Furthermore, $w_s^r$, is a binary variable, which takes 1 value whether vehicle $r$ enters the cluster $s$, and 0 in other case. Finally, the binary variable $x_{ij}^r$ is 1 if the vehicle $r$ uses the arc $(i, j)$, and 0 otherwise.

The main problem is now to minimize:
\begin{equation}
	 \sum_{i = 0}^{n} \ \sum_{j = 0}^{n} \ \sum_{r = 1}^{k} {d_{ij}x^r_{ij}} \label{ACVRPSPDVCFPeq1}
\end{equation}
where
\begin{equation}
	x^r_{ij} \in \{0,1\}, \ \ \ i,j = 0, \dots ,n; i \neq j; r = 1 \dots k,  \label{ACVRPSPDVCFPeq2}
\end{equation}
\begin{equation}
	w^r_s \in \{0,1\}, \ \ \ r = 1, \dots ,k; s = 1,...,c, \label{ACVRPSPDVCFPeq1_1}
\end{equation}
\begin{equation}
	y_{ij} \geq 0, \ \ \ i,j = 0, \dots ,n, \label{ACVRPSPDVCFPeq1_2}
\end{equation}
\begin{equation}
	z_{ij} \geq 0, \ \ \ i,j = 0, \dots ,n.  \label{ACVRPSPDVCFPeq1_3}
\end{equation}
This is subject to the following constraints:
%Primero las que tienen que ver con la naturaleza VRP
\begin{equation}
	\sum_{i = 0}^{n} \ \sum_{r = 1}^{k} {x^r_{ij}=1}, \ \ \ j = 0, \dots ,n; i \neq j, \label{ACVRPSPDVCFPeq3}
\end{equation}
\begin{equation}
	\sum_{j = 0}^{n} \ \sum_{r = 1}^{k} {x^r_{ij}=1}, \ \ \ i = 0, \dots ,n; j \neq i,  \label{ACVRPSPDVCFPeq4}
\end{equation}
\begin{equation}
	\sum_{j = 0}^{n} \ \sum_{r = 1}^{k} {x^r_{0j}=k}, \label{ACVRPSPDVCFPeq5}
\end{equation}
\begin{equation}
	\sum_{i = 0}^{n} \ \sum_{r = 1}^{k} {x^r_{i0}=k}, \label{ACVRPSPDVCFPeq6}
\end{equation}
\begin{equation}
	\sum_{i = 0}^{n} {x^r_{ij}} - \sum_{l = 0}^{n} {x^r_{jl}} = 0, \ j = 0, \dots ,n; r = 1 \dots k, \label{ACVRPSPDVCFPeq7}
\end{equation}
\begin{equation}
	\sum_{j = 0}^{n} {x^r_{ij}} - \sum_{l = 0}^{n} {x^r_{li}} = 0, \ i = 0, \dots ,n; r = 1 \dots k, \label{ACVRPSPDVCFPeq8}
\end{equation}

%Ahora las que tienen que ver con la naturaleza del simultaneous pickup and delivery
\begin{equation}
	\sum_{i = 0}^{n} {y_{ji}} - \sum_{i = 0}^{n} {y_{ij}} = p_j, \ \ \ j = 0, \dots ,n, \label{ACVRPSPDVCFPeq9}
\end{equation}
\begin{equation}
	\sum_{i = 0}^{n} {z_{ji}} - \sum_{i = 0}^{n} {z_{ij}} = d_j, \ \ \ j = 0, \dots ,n,  \label{ACVRPSPDVCFPeq10}
\end{equation}
\begin{equation}
	y_{ij} + z_{ij} \leq Q \sum_{r = 1}^{k} {x^r_{ij}}, \ \ \ i,j = 0, ..., n,  \label{ACVRPSPDVCFPeq11}
\end{equation}

%Ahora las que tienen que ver con la naturaleza de los caminos prohibidos

\begin{equation}
	\sum_{i = 0}^{n} \ \sum_{r = 1}^{k} {d_{ij}x^r_{ij}<\infty}, \ \ \ j = 0, \dots ,n; i \neq j, \label{ACVRPSPDVCFPeq12}
\end{equation}
\begin{equation}
	\sum_{j = 0}^{n} \ \sum_{r = 1}^{k} {d_{ij}x^r_{ij}<\infty}, \ \ \ i = 0, \dots ,n; j \neq i,  \label{ACVRPSPDVCFPeq13}
\end{equation}

%Ahora las que tienen que ver con la clusterización
\begin{equation}
\sum_{r = 1}^{k} {w^k_{s}} = 1 \ \ \ s = 1,...,c.  \label{ACVRPSPDVCFPeq14}
\end{equation}

The first clause represents the objective function, which is the sum of the costs of all routes of the solution, and it must be minimized. The formulas (\ref{ACVRPSPDVCFPeq2}), (\ref{ACVRPSPDVCFPeq1_1}), (\ref{ACVRPSPDVCFPeq1_2}) and (\ref{ACVRPSPDVCFPeq1_3}) depict the nature of the variables $x_{ij}^r$, $w^r_s$, $y_{ij}$ and $z_{ij}$. Equations (\ref{ACVRPSPDVCFPeq3}) and (\ref{ACVRPSPDVCFPeq4}) assure that all the nodes are visited exactly once. On the other hand, constraints (\ref{ACVRPSPDVCFPeq5}) and (\ref{ACVRPSPDVCFPeq6}) ensure that the total amount of vehicles leaving the depot, and the number of vehicles that return to it is the same. Besides, the correct flow of each route is ensured by functions (\ref{ACVRPSPDVCFPeq7}) and (\ref{ACVRPSPDVCFPeq8}).

Additionally, restrictions (\ref{ACVRPSPDVCFPeq9}) and (\ref{ACVRPSPDVCFPeq10}) guarantee that the flows for the collection and delivery demands, respectively, are properly conducted. These formulas ensure that both demands are satisfied for every customer. Furthermore, constraint (\ref{ACVRPSPDVCFPeq11}) assures that the total capacity of any vehicle is never exceeded; it also establishes that pick-up and delivery demands will only be transported using arcs included in the solution \cite{montane2006tabu}.

On the other hand, functions (\ref{ACVRPSPDVCFPeq12}) and (\ref{ACVRPSPDVCFPeq13}) ensure that every trip between one node and another has not an infinite cost. Thus, it is guaranteed that forbidden paths will not form part of the final solution. Finally, restriction (\ref{ACVRPSPDVCFPeq14}) assures that only one vehicle enters every cluster. This function, together with the above mentioned (\ref{ACVRPSPDVCFPeq3}) and (\ref{ACVRPSPDVCFPeq4}), ensures that all the customers belonging the same cluster are visited by the same vehicle.

\section{Firefly algorithm}
\label{sec:firefly}

As has been stated in the introduction, a discrete firefly algorithm (DFA) is proposed in this work to address the designed AC-VRP-SPDVCFP. In this section, the description of the classic FA is shown first (Section \ref{sec:classicFirefly}). Then, the proposed DFA is described in detail in Section \ref{sec:discreteFirefly}.

\subsection{Classic Firefly Algorithm}
\label{sec:classicFirefly}

The basic FA was first developed by Xin-She Yang in 2008 \cite{yang2008nature,yang2009firefly}, and it was based on the idealized behaviour of the flashing characteristics of fireflies. To properly understand the algorithm, it is important to highlight the following three idealized rules \cite{yang2008nature}:

\begin{itemize}
	\item All the fireflies of the swarm are unisex, and one firefly will be attracted to other ones regardless of their sex.
\end{itemize}

\begin{itemize}
	\item Attractiveness is proportional to the brightness, which means that, for any two fireflies, the brighter one will attract the less bright one. The attractiveness decreases as the distance between the fireflies increases. Furthermore, if one firefly is the brightest one of the swarm, it moves randomly.
\end{itemize}

\begin{itemize}
	\item The brightness of a firefly is directly determined by the objective function of the problem under consideration. In this manner, for a maximization problem, the brightness can be proportional to the objective function value. On the other hand, for a minimization problem, it can be the reciprocal of the objective function value.
\end{itemize}

\begin{algorithm}[tb]
	 \SetAlgoLined
		Define the objective function $f(x)$\;
		Initialize the firefly population $X = {x_1,x_2,...,x_n}$\;
		Define the light absorption coefficient $\gamma$\;
		\For{each firefly $x_i$ in the population}{
				Initialize light intensity $I_i$\;
		}
		\Repeat{termination criterion reached}{
				\For{each firefly $x_i$ in the swarm}{
					\For{each other firefly $x_j$ in the swarm}{
						\If{$I_j > I_i$}{
							Move firelfy $x_i$ toward $x_j$ \;
						}
						Attractiveness varies with distance $r$ via exp(-$\gamma$$r$)\;
						Evaluate new solutions and update light intensity\;
				}}
		Rank the fireflies and find the current best\;
		}
		Rank the fireflies and return the best one\;
   \caption{Pseudocode of the basic FA.}
	 \label{alg:FA}
\end{algorithm}

In Algorithm \ref{alg:FA}, the pseudocode of the basic FA is shown, which was proposed by Yang in \cite{yang2008nature}. In line with this, there are three important factors to consider in the FA: the attractiveness, the distance and the movement. In the basic version of the FA these factors are tackled in the following way. First of all, the attractiveness of a firefly is determined by its light intensity, and it can be calculated using this formula:

\begin{equation}
	\beta(r) = \beta_0 e^{-\gamma r^2}   \label{FA1}
\end{equation}

On the other hand, in the basic FA the distance $r_{ij}$ between any two fireflies $i$ and $j$ is calculated using the Cartesian distance, and it is computed by the following equation:
\begin{equation}
	 r_{ij} = || X_i - X_j || = \sqrt{\sum_{k=1}^d (X_{i,k} - X_{j,k})^2} \label{FA2}
\end{equation}
where $X_{i,k}$ is the $k$th component of the spatial coordinate $X_i$ of the $i$th firefly. Finally, the movement of a firefly $i$ toward any other brighter firefly $j$ is determined by this formula:

\begin{equation}
	 X_i = X_i + \beta_0 e^{-\gamma r^2_{ij}}(X_j - X_i) + \alpha (rand-0.5) \label{FA3}
\end{equation}
where $\alpha$ is the randomization parameter and $rand$ is a random number uniformly distributed in [0,1]. On the other hand, the second term of the equation stems from the attraction assumption.

\subsection{The proposed Discrete Firefly Algorithm}
\label{sec:discreteFirefly}

It is noteworthy that the original FA was developed primarily for solving continuous optimization problems. For this reason, the classic FA cannot be applied directly to solve the proposed AC-VRP-SPDVCFP. Therefore, some modifications of the original FA are needed in order to prepare it for addressing such AC-VRP-SPDVCFP problem.

First of all, in the proposed DFA, each firefly in the swarm represents a possible and feasible solution for the AC-VRP-SPDVCFP. All the fireflies are initialized randomly. Additionally, as has been detailed in Section \ref{sec:VRP}, the sum of the costs of all routes of the solution has been used as the objective function. Therefore, the AC-VRP-SPDVCFP is a minimization problem, in which the fireflies with a lower objective function value are the most attractive ones. In addition, the concept of light absorption is also represented in this version of the FA. In this case, $\gamma$ = 0.95, and this parameter is used in the way as can be seen in Equ.~(\ref{FA3}). This parameter has been set following the guidelines proposed in several studies of the literature \cite{yang2009firefly,yang2008nature}.

Besides that, the distance between two different fireflies is represented by the Hamming Distance. The Hamming distance between two fireflies is the number of non-corresponding elements in the sequence. In the proposed AC-VRP-SPDVCFP, the comparison is made cluster by cluster. For example, taking into account two random fireflies, and one random cluster $c$ composed by 8 nodes:
\[x_1 (\textrm{cluster-} c): \{0,1,2,3,4,5,6,7\}, \]
\[x_2 (\textrm{cluster-} c): \{0,1,3,2,5,4,6,7\}, \]
the Hamming Distance between $x_1$ and $x_2$ for the cluster $k$ would be 4. This same comparison is made for every cluster. In this way, the total distance between fireflies $i$ and $j$ is the sum of all the distances for every cluster.

Finally, the movement of a firefly $i$ attracted to another brighter firefly $j$ is determined by
\begin{equation}
	 n = \textrm{Random}(2, r_{ij} \cdot \gamma^g) \label{FA4}
\end{equation}

\begin{equation}
	 x_i = \textrm{InsertionFunction}(x_i, n) \label{FA5}
\end{equation}
where $r_{ij}$ is the Hamming Distance between firefly $i$ and firefly $j$, and $g$ is the iteration number. In this case, the length of the movement of a firefly will be a random number between 2 and $r_{ij}$ $\cdot$ $\gamma^g$. As for the movement function, the Insertion Function has been used. This function selects and extracts one random node from a random route. After that, this node is re-inserted in a random position inside the cluster of the selected node. This function takes into account the capacity constraint, in order not to create infeasible solutions.

Following the same philosophy as other previously developed and published FA for the Traveling Salesman Problem \cite{jati2011evolutionary,zhou2014multi}, fireflies in the proposed DFA do not have directions to move. Instead, fireflies move using evolution strategies. In this way, each firefly moves using $n$ times the Insertion Function, generating $n$ potential successors. After these $n$ movements, the best one is performed, generating the new firefly.

The pseudocode of the presented DFA is shown in Algorithm \ref{sec:exp}. In lines 1-3 the initialization phase of the algorithm is carried out, in which fireflies are initialized and evaluated. Besides, the $\gamma$ parameter is initialized as previously described. In addition, in lines 10-12 the movement process is performed. In line 10 the distance between the selected $x_i$ and $x_j$ is calculated via Hamming Distance. Once the distance is obtained, the $n$ parameter is calculated, which is a random number between 2 and $r_{ij} \cdot \gamma^g$. Finally, the movement is performed in line 12 using the Insertion Function as explained before. After this movement process, fireflies are evaluated in line 14 and ranked in line 17. This iterative procedure is repeated until termination criterion is reached.

\begin{algorithm}[tb]
	 \SetAlgoLined
		Initialize the firefly population $X = {x_1,x_2,...,x_n}$\;
		$\gamma$ = 0.95\;
		\For{each firefly $x_i$ in the population}{
				$I_i$\ = $fitness(x_i)$\;
		}
		\Repeat{termination criterion reached}{
				\For{each firefly $x_i$ in the swarm}{
					\For{each other firefly $x_j$ in the swarm}{
						\If{$I_j < I_i$}{
							$r_{ij}$ = HammingDistance($x_i$,$x_j$)\;
							$n$ = Random(2,$r_{ij}$ $\cdot$ $\gamma^g$)\;
							$x_i$ = InsertionFunction ($x_i$,$n$)\;
						}
						Evaluate new solutions and update light intensity\;
				}}
		Rank the fireflies and find the current best\;
		}
		Rank the fireflies and return the best one\;
   \caption{Pseudocode of the proposed DFA.}
	 \label{alg:DFA}
\end{algorithm}

\section{Experimentation}
\label{sec:exp}

The computational experiments carried out in this study are described in this section. First of all, the details of the proposed benchmark for the AC-VRP-SPDVCFP are detailed (Section \ref{sec:benchmark}). After that, the results obtained by the developed DFA for this benchmark are presented (Section \ref{sec:results}). In order to prove that the DFA is a promising metaheuristic for solving routing problems, the results obtained by the DFA have been compared with the ones obtained by the EA and the ESA. In addition, two different statistical tests have been conducted, with the aim of obtaining rigorous and fair conclusions (Section \ref{sec:disc}).

\subsection{The proposed benchmark for the VRP}
\label{sec:benchmark}

\begin{figure}[tb]
	\centering
		\includegraphics[width=0.45\textwidth]{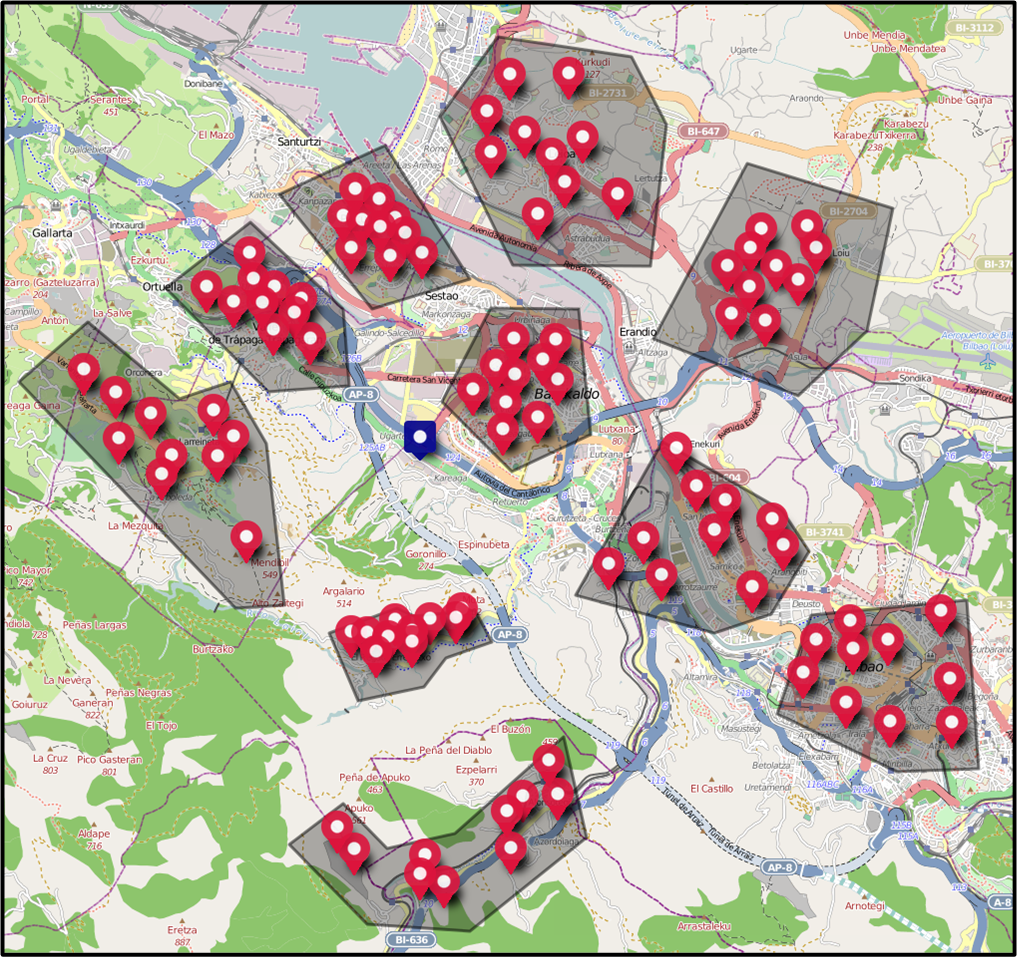}
	\caption{Geographical locations of the depot, customers and clusters around the province of Bizkaia. Source: Open Street Maps, via uMap, accessed Sept 2015.}
	\label{fig:map}
\end{figure}

The type of R-VRP proposed in this paper has never been treated before in the literature. It is for this reason that there is no benchmark available in the literature for the AC-VRP-SPDVCFP. In this work, a benchmark composed by 15 instances is proposed. At the same time, these instances are composed of 50 to 100 nodes. Every node represents a customer, and all nodes are placed in real geographical locations, which are located in the province of Bizkaia, Spain. In addition, the maximum number of clusters has been established as 10, existing also instances with 5, and 8 of them. In the Figure \ref{fig:map}, a map with the geographical locations of the depot, the customers and the clusters are shown.
This map has been made using Open Streep Maps technology, via uMap tool\footnote{http://umap.openstreetmap.fr}.

The clusters have been organized in order of appearance. That is, the nodes 1-10 compose cluster 1, nodes 11-20 for cluster 2, and so on. It is important to highlight that every cluster has the same number of customers in all instances. Besides that, as has been pointed out in Section \ref{sec:generalDesc}, each customer has assigned two kinds of demands: one associated with the delivery $d_i$ and the other with the pick-up $p_i$. The assignments of these demands
have been performed following this procedure:

\begin{equation}
	  d_i = 10, \; p_i = 5, \ \ \ \forall i \in \{1, 5, 9, \dots , 97\}, \label{demands1}
\end{equation}
\begin{equation}
	  d_i = 10, \; p_i = 0, \ \ \ \forall i \in \{2, 6, 10, \dots , 98\}, \label{demands2}
\end{equation}
\begin{equation}
	  d_i = 5, \; p_i = 3, \ \ \ \forall i \in \{3, 7, 11, \dots , 99\}, \label{demands3}
\end{equation}
\begin{equation}
	  d_i = 5, \; p_i = 0, \ \ \ \forall i \in \{4, 8, 12, \dots , 100\}, \label{demands4}
\end{equation}
where $d_0$=0 and $p_0$=0, taking into account that $v_0$ is considered the depot. In addition, the costs of traveling from any customer $i$ to other customer $j$ have been established following the procedure depicted in Algorithm \ref{alg:distances}. By this method, the asymmetry characteristic is met. It is important to highlight that these costs are assigned for the ``off-peak" period. These costs are incremented when they are conducted on the ``peak" period, following the procedure shown in Algorithm \ref{alg:distances2}. In is noteworthy that, with the intention of simplifying the complexity of the problem, the traveling time between any node $i$ and $j$ is the same as its traveling cost (in seconds).

\begin{algorithm}[tb]
	 \SetAlgoLined
		\For{$\forall i \in \{1,2, \dots , 99\}$}{
			\For{$\forall j \in \{i+1, \dots , 100\}$}{
				
				$d_{ij}$ = \textrm{EuclideanDistance}(i,j)\;
				
				\eIf{$j$ \textrm{is an odd number}}{
					$d_{ji}$ = \textrm{EuclideanDistance}(j,i) $\cdot$ 1.2 \;
				}{
				  $d_{ji}$ = \textrm{EuclideanDistance}(j,i) $\cdot$ 0.8 \;
				}
		}}
   \caption{Procedure of travel costs assignment for ``off-peak" period.}
	 \label{alg:distances}
\end{algorithm}

\begin{algorithm}[tb]
	 \SetAlgoLined
		\For{$\forall i \in \{1,2, \dots , 99\}$}{
			\For{$\forall j \in \{i+1, \dots , 100\}$}{
				
				$d_{ij}$ = \textrm{EuclideanDistance} $\cdot$ 1.3\;
				\eIf{$j$ \textrm{is an odd number}}{
					$d_{ji}$ = (\textrm{EuclideanDistance}(j,i) $\cdot$ 1.2)  $\cdot$ 1.2 \;
				}{
				  $d_{ji}$ = (\textrm{EuclideanDistance}(j,i) $\cdot$ 0.8) $\cdot$ 1.4 \;
				}
		}}
   \caption{Procedure of travel costs assignment for ``peak" period.}
	 \label{alg:distances2}
\end{algorithm}

Finally, depending on the instance, some paths are chosen in each cluster to be forbidden. In Table \ref{tab:instances}, the characteristics of all the instances developed for the benchmark are summarized. In order to understand the content of this table correctly, the following clarifications should be made: Osaba\_50\_1\_1 and Osaba\_50\_1\_2 are comprised by 5 clusters, which are the clusters \{1, 3, 5, 7, 9\}. On the other hand, Osaba\_50\_2\_1 and Osaba\_50\_2\_2 are made up by clusters \{2, 4, 6, 8, 10\}. Moreover, clusters in Osaba\_50\_1\_3 and Osaba\_50\_1\_4 are composed by 5 nodes. In this case, this customers are the first 5 of every cluster. This is in contrast with what happens in Osaba\_50\_2\_3 and Osaba\_50\_2\_4, where the 10 clusters are comprised of the last 5 clients of every of them. Lastly, to complete all Osaba\_80\_X instances, the first 8 clusters, or nodes (depending on the case) have been chosen.With the aim of allowing the replication of this experimentation, it is noteworthy that the benchmark developed is available under request to the corresponding author of this paper, or via Web\footnote{http://research.mobility.deustotech.eu/media/publication \_resources/Instances\_Osaba\_AC-VRP-SPDVCFP.rar.}

\begin{table}[t]
	\centering
	\setlength{\tabcolsep}{2pt}
	\caption{Summary of the benchmark proposed for the AC-VRP-SPDVCFP. \textit{Forbidden paths} depicts the number of forbidden arcs in each cluster.}
	\scalebox{0.8}{
		\begin{tabular}{| l || r | r | r | r |}
			\hline Instance & Nodes & Clusters & Vehic. capacity & Forbidden paths \\
			\hline Osaba\_50\_1\_1 & 50 & 5 & 240 & 5\\
			Osaba\_50\_1\_2 & 50 & 5 & 160 & 10 \\
			Osaba\_50\_1\_3 & 50 & 10 & 240 & 5 \\
			Osaba\_50\_1\_4 & 50 & 10 & 160 & 10 \\
			\hline
			Osaba\_50\_2\_1 & 50 & 5 & 240 & 5\\
			Osaba\_50\_2\_2 & 50 & 5 & 160 & 10 \\
			Osaba\_50\_2\_3 & 50 & 10 & 240 & 5 \\
			Osaba\_50\_2\_4 & 50 & 10 & 160 & 10 \\
			\hline
			Osaba\_80\_1 & 80 & 8 & 240 & 5 \\
			Osaba\_80\_2 & 80 & 8 & 160 & 10 \\
			Osaba\_80\_3 & 80 & 10 & 240 & 5 \\
			Osaba\_80\_4 & 80 & 10 & 160 & 10 \\
			\hline
			Osaba\_100\_1 & 100 & 10 & 140 & 5 \\
			Osaba\_100\_2 & 100 & 10 & 260 & 10 \\
			Osaba\_100\_3 & 100 & 10 & 320 & 10 \\
			\hline
		\end{tabular}
	}
	\label{tab:instances}
\end{table}

\subsection{Results}
\label{sec:results}

All the tests conducted in this work have been performed on an Intel Core i5 – 2410 laptop, with 2.30 GHz and a RAM of 4 GB. Java has been used as the programming language. All the 15 instances described in the previous section have been used in this experiment. Every instance has been run 20 times. As has been said before, the results obtained by the DFA are compared with the ones obtained by the EA and the ESA. The reason why these two techniques have been used for this experimentation can be summarized as follows: First of all, both meta-heuristics are well-known, and they have been frequently used to solve routing problems. Proving that the DFA outperforms these two techniques can be concluded that it is a promising technique to solve R-VRPs. On the other hand, all these three techniques have two similarities: all of them base the movement of their individuals on a short-step operator, and they are easy to implement and can be adapted to solve new problems.

\begin{table}[t]
	\centering
	\setlength{\tabcolsep}{2pt}
	\caption{Parametrization of the EA and ESA for the proposed AC-VRP-SPDVCFP, where $-sup \Delta f$ is the difference in the objective function of the best and the worse individuals of the initial population, and $p$=0.95.}
	\scalebox{0.7}{
		\begin{tabular}{| l | l || l | l |}
		  \hline
		  \multicolumn{2}{|c||}{EA} & \multicolumn{2}{c|}{ESA}\\
			\hline
			\hline Parameter & Value & Parameter & Value \\
			\hline
			Population size & 100 & Population size & 100 \\
			Mutation functions & Insertion Function & Successor Function & Insertion Function \\
			Mutation prob. & 1.0  & Temperature & $-sup \Delta f/ ln(p) $\\
		  Survivor func. & 70\% Elitist - 30\% Random  & Cooling constant & 0.95 \\
			\hline
			
			\hline
		\end{tabular}
	}
	\label{tab:Parametrization2}
\end{table}

It is worth highlighting that, as far as possible, the same operators and similar parameters have been used for all the algorithms implemented for the experimentation. In this way, the intention is to conclude which algorithm obtains better results using similar operators and a similar number of times. Furthermore, with the intention of facilitating the replicability of this study, the parameters used for the EA and ESA are shown in Table \ref{tab:Parametrization2}. It is worth pointing out that all the individuals are randomly generated. Additionally, as for the termination criterion, every algorithm finishes its execution when there are $n+\sum_{k=1}^n{k}$ function evaluations without improvements in the best solution, where $n$ is the size of the problem. In addition, the parameters used for DFA are the described in Section \ref{sec:discreteFirefly}. In case of DFA, a population of 100 fireflies has also been used.

Finally, in this study, the permutation codification has been used for the representations of the solutions. Thus, each solution $X$ is encoded by an unique permutation of numbers, which represents the different routes that compose that solution. Besides that, with the aim of distinguishing the routes in one solution, they are separated by zeros.

Before starting the tests, a small study about the parametrization of the DFA is shown. It should be taken into account that performing a comprehensive study on the parameterization of the algorithm would be very extensive. That study has been planned as future work, since the objective of this paper is to present the NDSRP model, and to demonstrate that the DFA shows an adequate performance applied to an R-VRP. For this reason, in this paper a small portion of that study is shown, in order to justify the population size used. For that, we compare 4 different versions of the DFA, each one with a different population size. Each version is called $DFA_x$, where $x$ is the popualtion size. For this small test, 4 different instances have been used. In Table \ref{tab:param} the results of this experimentation are depicted. In this table, the results average (avg.), and average runtime (Time, in seconds) are shown. In addition, the last row of the table represents the average ranking for each alternative.

\begin{table}[th]
	\centering
	\setlength{\tabcolsep}{2pt}
	\caption{Small study about the population size on the DFA}
	\scalebox{0.75}{
		\begin{tabular}{| l || r r | r r | r r | r r |}
			\hline
		  \multicolumn{1}{|c||}{Instance} & \multicolumn{2}{c||}{$DFA_{25}$} & \multicolumn{2}{c||}{$DFA_{50}$} & \multicolumn{2}{c|}{$DFA_{100}$}& \multicolumn{2}{c|}{$DFA_{150}$}\\
			\hline
			\hline Name & Avg. & Time & Avg. & Time & Avg. & Time & Avg. & Time\\
			\hline Osaba\_50\_1\_1 & 51945.7 & 14.8 & 51561.3 & 25.8 & 50989.5 & 37.9  & \textbf{50934.3} & 68.9\\
			Osaba\_50\_1\_2 & 57398.7 & 15.8 & 56721.8 & 26.9 & \textbf{56203.8} & 35.1  & 56213.7 & 71.8 \\
			Osaba\_80\_3 & 92990.39 & 33.7 &  91663.8 & 47.3 & \textbf{89512.0} & 75.3  & 89531.0 & 112.3 \\
			Osaba\_100\_1 & 110206.94 & 51.4 & 108241.6 & 81.8  & 107799.5 & 112.8  & \textbf{107745.7} & 176.4 \\
			\hline
			\hline
			Ranking & \multicolumn{2}{c|}{4} & \multicolumn{2}{c|}{3} & \multicolumn{2}{c|}{1.5} & \multicolumn{2}{c|}{1.5} \\
			\hline
		\end{tabular}
	}
	\label{tab:param}
\end{table}

Some conclusions can be drawn if the data presented in Table \ref{tab:param} are analyzed. It can be seen a slight trend of improvement in the results when the population size increases. Even so, this fact involves a significant increase of runtime, which is not directly proportional to the improvement in results. In this paper, to achieve the proposed, the option which best balances the runtime and results quality has been selected. This option is $DFA_{100}$. This version has acceptable execution times, and is the alternative that more improvement offers regarding its previous version. $DFA_{150}$, on the other hand, needs very high execution times without offering significant improvements in the results quality.

Once carried out this small study, the results obtained by the three techniques for the proposed benchmark are summarized in Table \ref{tab:Results}. In this table, the results average (avg.), standard deviation (S. dev), average runtime (Time, in seconds) and average convergence time (C. T, in seconds) are shown. Additionally, the best results averages have been represented bolded. Besides that, the best results found for each instance are represented in Table \ref{tab:BestResults}. This table also shows the number of vehicles needed to perform every solution, and the technique which reached each of these results. Since this is the first appearance of the AC-VRP-SPDVCFP in the literature, these solutions are considered the best ones found until the publication of this paper.

\begin{table*}[t]
	\centering
	\setlength{\tabcolsep}{2pt}
	\caption{Results of DFA, ESA and EA for the proposed AC-VRP-SPDVCFP.}
	\scalebox{0.9}{
		\begin{tabular}{| l || r r r r || r r r r || r r r r|}
		  \hline
		  \multicolumn{1}{|c||}{Instance} & \multicolumn{4}{c||}{DFA} & \multicolumn{4}{c||}{ESA} & \multicolumn{4}{c|}{EA}\\
			\hline
			\hline Name & Avg. & S. dev. & Time & C. T. & Avg. & S. dev. & Time & C. T. & Avg. & S. dev. & Time & C. T. \\
			\hline Osaba\_50\_1\_1 & \textbf{50989.5} & 234.9 & 37.9 & 26.2 &  51632.6 & 479.0 & 31.0 & 22.7 &  51569.3 & 649.7 & 38.4 & 26.8 \\
			Osaba\_50\_1\_2 & \textbf{56203.8} & 203.4 & 35.1 & 25.4 &  56929.4 & 459.1 &  31.7 & 21.0 & 57008.8 & 540.9 & 34.0  & 26.4 \\
			Osaba\_50\_1\_3 & \textbf{71730.0} & 1486.2 & 37.6 & 22.3 & 72298.3 & 1484.4 & 32.1 & 17.4 & 72490.7 & 1413.4 & 36.1  & 21.8 \\
			Osaba\_50\_1\_4 & \textbf{78883.8} & 1193.4 & 35.7 & 23.9 & 79168.0 & 1788.8 & 29.4 & 20.0 & 79207.9 & 1612.4 & 34.8 & 24.5 \\
			\hline
			Osaba\_50\_2\_1 & \textbf{49276.0} & 392.4 & 38.6 & 22.3 & 49751.3 & 538.3 & 33.0 & 19.8 & 49416.8 & 760.6 & 36.9  & 20.4 \\
			Osaba\_50\_2\_2 & \textbf{54589.6} & 628.1 & 37.8 & 28.5 & 54871.6 & 621.0 &  31.8 & 23.7 & 55007.3 & 837.0 & 38.1  & 27.9 \\
			Osaba\_50\_2\_3 & \textbf{69631.4} & 2400.1 & 38.4 & 25.9 & 71352.6 & 2883.4 & 33.6 &  21.6 & 71286.3 & 3037.6 & 37.5  & 25.1 \\
			Osaba\_50\_2\_4 & \textbf{80543.7} & 1512.3 & 36.1 & 22.6 & 81122.0 & 1675.3 & 32.4 & 19.1 & 81297.3 & 1894.3 & 36.0 & 23.1 \\
			\hline
			Osaba\_80\_1 & 82307.8 & 1043.4 & 72.4 & 57.6 & \textbf{80834.2} & 1801.4 & 67.0 & 53.4 & 81779.6 & 2018.4 & 71.8 & 58.0\\
			Osaba\_80\_2 & \textbf{89324.9} & 698.0 & 74.0 & 60.7 & 89989.6 & 1134.4 &  67.6 & 54.3  & 90090.6 & 1032.5 & 73.4 & 59.4 \\
			Osaba\_80\_3 & 89512.0 & 1414.0 & 75.3 & 60.4 & \textbf{89431.2} & 2680.1 & 68.5 & 55.2 & 89883.1 & 2949.6 & 75.6 & 59.8 \\
			Osaba\_80\_4 & \textbf{104601.9} & 1299.2 & 75.0 & 64.8 & 105141.7 & 1963.3 &  69.4 & 61.3 & 106689.3 & 1832.9 & 74.8 & 65.3\\
			\hline
			Osaba\_100\_1 & \textbf{107799.5} & 1501.8 & 136.1 & 112.8 & 109183.6 & 1540.4 & 129.3 & 108.1 & 109614.4 & 1700.7 & 140.7 & 116.7 \\
			Osaba\_100\_2 & \textbf{100522.9} & 1683.0 & 138.0 & 109.2 & 101708.0 & 1644.1 & 130.4 & 102.0 & 101908.1 & 1699.0 & 141.0 & 111.9 \\
			Osaba\_100\ 3 & \textbf{95553.7} & 1470.6 & 139.7 & 105.3 & 95641.3 & 2687.3 & 128.0 & 98.4 & 95893.7 & 2472.0 & 140.3 & 108.4\\
			\hline
		\end{tabular}
	}
	\label{tab:Results}
\end{table*}

\begin{table}[th]
	\centering
	\setlength{\tabcolsep}{2pt}
	\caption{Best solutions found for each instance of the proposed benchmark.}
	\scalebox{0.8}{
		\begin{tabular}{| l || r | r | r |}
			\hline Name & Best Result & Vehicles & Technique\\
			\hline Osaba\_50\_1\_1 & 50641.46 & 2 & DFA \\
			Osaba\_50\_1\_2 & 55923.48 & 3 & DFA  \\
			Osaba\_50\_1\_3 & 68535.19 & 2 & DFA  \\
			Osaba\_50\_1\_4 & 76276.83 & 3 & DFA \\
			\hline
			\hline
			Osaba\_50\_2\_1 & 48819.71 & 2 & DFA  \\
			Osaba\_50\_2\_2 & 53876.86 & 3 & DFA  \\
			Osaba\_50\_2\_3 & 65059.03 & 2 & DFA  \\
			Osaba\_50\_2\_4 & 77497.06 & 3 & DFA \\
			\hline
			\hline
			Osaba\_80\_1 & 77660.84 & 3 & ESA \\
			Osaba\_80\_2 & 87835.50 & 4 &  DFA\\
			Osaba\_80\_3 & 83713.72 & 3 & ESA \\
			Osaba\_80\_4 & 101497.11 & 5 & DFA \\
			\hline
			\hline
			Osaba\_100\_1 & 105165.35 & 5 &  DFA\\
			Osaba\_100\_2 & 97924.51 & 4 & DFA\\
			Osaba\_100\ 3 & 88966.48 & 3 & ESA\\
			\hline
		\end{tabular}
	}
	\label{tab:BestResults}
\end{table}

\subsection{Analysis and Discussion}
\label{sec:disc}

\begin{figure*}[tb]
	\centering
		\includegraphics[width=0.5\textwidth]{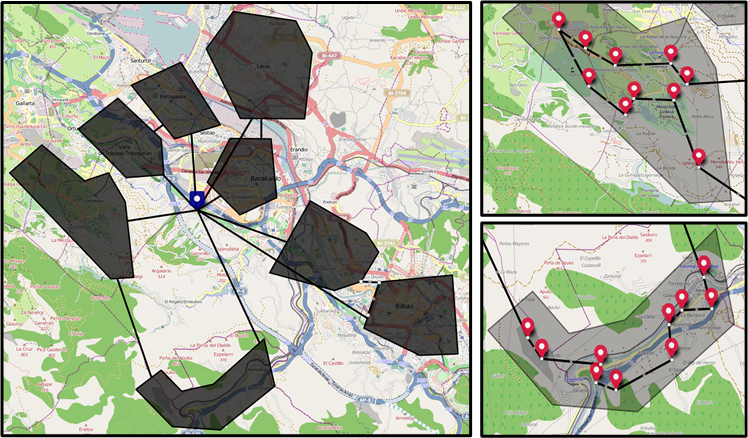}
	\caption{Partial representation of the best solution found by the DFA for the instance Osaba\_80\_2. Source: Open Street Maps, via uMap, accessed Sept 2015.}
	\label{fig:map2}
\end{figure*}

\begin{figure*}[tb]
	\centering
		\includegraphics[width=0.5\textwidth]{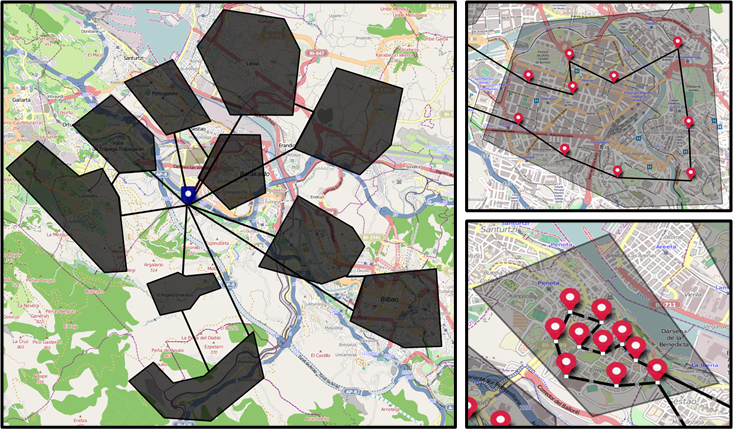}
	\caption{Partial representation of the best solution found by the DFA for the instance Osaba\_100\_1. Source: Open Street Maps, via uMap, accessed Sept 2015.}
	\label{fig:map3}
\end{figure*}

Analyzing the results in Table \ref{tab:Results}, the first conclusion that can be drawn is the following one: DFA outperforms clearly the other algorithms in terms of results. Specifically, the DFA performs better than ESA in 86.66\% of the instance (13 out of 15), and in 93.33\% of the cases comparing with the EA (14 out of 15). Additionally, the supremacy of DFA can also be seen in Table \ref{tab:BestResults}, having obtained the best solution in 80\% of the instances (12 out of 15). In relation with the data shown in Table \ref{tab:BestResults}, in Figure \ref{fig:map2} and Figure \ref{fig:map3}, the partial representation of the best solutions found by the DFA for the instances Osaba\_80\_2 and Osaba\_100\_1 are shown. It is noteworthy that, due to the complex nature of the problem, the complete representation of these solutions would lead to maps with some overlapping lines, complicating the visibility of the whole solution. Therefore, these solutions are shown at cluster-level, showing also several clusters in detail.

Another important factor that is worth mentioning is the robustness of the DFA in relation to the other techniques. It should be clarified that the robustness is the capacity of a technique to obtain similar solutions in every run. As can be seen in Table \ref{tab:Results}, the standard deviation of the results obtained by the DFA is lower than the ones presented by the other metaheuristics in most instances (12 out of 15). This means that the quality of the solutions provided by the DFA move in a narrow range. This characteristic gives robustness and reliability to the algorithm, something crucial if the technique is applied a real environment.

Regarding the runtimes, in Table \ref{tab:Results} it can be seen how DFA and EA have similar execution times, while the ESA takes less time than its competitors. These same conclusions can be drawn looking at the convergence times, where ESA shows that it needs less time to reach the final solution.  This fact can be analyzed in the following way: The DFA needs more time to reach its final solution, showing a better exploration capacity than the ESA. On the other hand, it shows a better exploitation capacity than the EA, since it obtains better results needing similar execution and convergence times.

Besides that, two different statistical tests have been conducted in order to obtain rigorous and fair conclusions.  It is important to clarify that the result averages obtained by each technique have been use to perform these tests. The guidelines given by Derrac et al. in \cite{derrac2011practical} have been followed to perform this statistical analysis. First of all, the Friedman's non-parametric test for multiple comparisons has been used to check if there are any significant differences among all the techniques. The resulting Friedman statistic has been 17.73. Taking into account that the confidence interval has been stated at the 99\% confidence level, the critical point in a $\chi^2$ distribution with 2 degrees of freedom is 9.21. Since 17.73$>$9.21, it can be concluded that there are significant differences among the results reported by the three compared algorithms, being DFA the one with the lowest rank. Finally, regarding this Friedman's test, the computed p-value has been 0.000141.

To evaluate the statistical significance of the better performance of DFA, the Holm's post-hoc test has been conducted using DFA as control algorithm. The unadjusted and adjusted p-values obtained through the application of Holm's post-hoc procedure can be seen in Table \ref{tab:results_holms}. Analyzing this data, and taking into account that all the p-values are lower than 0.05, it can be concluded that DFA is significantly better than ESA and EA at a 95\% confidence level.

\begin{table}[tbh]
	\centering
	\caption{Average rankings returned by the Friedman's non-parametric test for DFA, ESA and EA.}
	\scalebox{0.7}{
		\begin{tabular}{|c|c|}\hline
			Algorithm&Average Ranking\\\hline
			DFA&1.2\\
			ESA&2.0667\\
			EA&2.7333\\
			\hline
		\end{tabular}
	}
	\label{tab:results_friedman}
\end{table}

\begin{table}[tbh]
	\centering
	\caption{Unadjusted and adjusted p-values obtained through the application of Holm's post-hoc procedure using DFA as control algorithm.}
	\scalebox{0.7}{
		\begin{tabular}{|c|c c|}\hline
		Algorithm & Unadjusted $p$& Adjusted $p$\\
		\hline
		ESA&0.017622&0.017622\\
		EA&0.000027&0.000054\\\hline
		\end{tabular}
	}
	\label{tab:results_holms}
\end{table}

\section{Conclusions and Further Work}
\label{sec:conc}
In this work, a new version of the newspaper delivery problem with recycling policy has been tackled. This problem has been modelled as a rich vehicle routing problem,  specifically, as an asymmetric and clustered vehicle routing problem with simultaneous pickup and deliveries, variable costs and forbidden paths. It is the first time that a problem like this is presented in the literature, for this reason, a benchmark composed by 15 instances has been also developed, using real-world geographical locations. To deal with such a complex problem, a discrete firefly algorithm has been developed. This application can be considered as the first application of a firefly algorithm to any rich vehicle routing problem. To prove that the proposed DFA is a promising technique, its performance has been compared with those obtained by two other well-known techniques: the evolutionary algorithm, and the evolutionary simulated annealing. The DFA has outperformed these two classic metaheuristics.

As for future work, it is intended to extend the application of the DFA to other complex real-world situations, related to transportation and logistics. In addition, comparison with more algorithms
and techniques will be carried out. Various improvements will be investigate so as to see if the results shown in this work for the AC-VRP-SPDVCFP can be improved. Besides this, it is intended to conduct a thorough study on the parameterization of the DFA (analyzing, for example, the time complexity). This study has not been done in this work because the main objective is to present the NDSRP model, and to demonstrate that the DFA shows an adequate performance applied to an R-VRP. Finally, it would be useful to perform AC-VRP-SPDVCFP instances using random changes in the asymmetric travel costs.

\section*{Acknowledgement}
This project was supported by the European Union’s Horizon 2020 research and innovation programme   through the TIMON: Enhanced real time services for optimized multimodal mobility relying on cooperative networks and open data project (636220). As well as by the projects TEC2013-45585-C2-2-R from the Spanish Ministry of Economy and Competitiveness, and PC2013-71A from the Basque Government.

\bibliographystyle{splncs}

\end{document}